\documentclass{article}

\usepackage{microtype}
\usepackage{graphicx}
\usepackage{subfigure}
\usepackage{booktabs} 
\usepackage{color,soul}  
\usepackage{hyperref}

\usepackage{multirow}
\usepackage[accepted]{icmlw2019generalization}
\icmltitlerunning{Utilizing Eye Gaze to Enhance the Generalization of Imitation Network to Unseen Environments}

\usepackage[english]{babel}

\newcommand{\placetextbox}[3]{
  \AddToShipoutPictureFG*{
    \put(\LenToUnit{#1\paperwidth},\LenToUnit{#2\paperheight}){\vtop{{\null}\parbox{\textwidth}{#3}}}%
  }%
}%

\begin{document}
\twocolumn[
\icmltitle{Utilizing Eye Gaze to Enhance the Generalization of Imitation Networks to Unseen Environments}



\icmlsetsymbol{equal}{*}

\begin{icmlauthorlist}
\icmlauthor{Congcong Liu}{equal,to}
\icmlauthor{Yuying Chen}{equal,to}
\icmlauthor{Lei Tai}{to}
\icmlauthor{Ming Liu}{to}
\icmlauthor{Bertram Shi}{to}
\end{icmlauthorlist}

\icmlaffiliation{to}{Department of Electronic and Computer Engineering, The Hong Kong University of Science and Technology, Hong Kong SAR, China}
\icmlcorrespondingauthor{Yuying Chen}{ychenco@connect.ust.hk}
\icmlcorrespondingauthor{Congcong Liu}{cliubh@connect.ust.hk}

\icmlkeywords{Machine Learning, ICML}

\vskip 0.3in
]

\printAffiliationsAndNotice{\icmlEqualContribution} 

\placetextbox{0.09}{1}{\small{{\textcopyright}  2019 IEEE. Personal use of this material is permitted. Permission from IEEE must be obtained for all other uses, in any current or future media, including reprinting/republishing this material for advertising or promotional purposes, creating new collective works, for resale or redistribution to servers or lists, or reuse of any copyrighted component of this work in other works.}}
\begin{abstract}
Vision-based autonomous driving through imitation learning mimics the behaviors of human drivers by training on pairs of data of raw driver-view images and actions.
However, there are other cues, e.g. gaze behavior, available from human drivers that have yet to be exploited.
Previous research has shown that novice human learners can benefit from observing experts' gaze patterns.
We show here that deep neural networks can also benefit from this. We demonstrate different approaches to integrating gaze information into imitation networks.
Our results show that the integration of gaze information improves the generalization performance of networks to unseen environments.
\end{abstract}

\section{Introduction}
\label{Introduction}
End-to-end deep learning has captured much attention and has been widely applied in many autonomous control systems.
Successful attempts have been made in end-to-end driving through imitation learning \cite{bojarski2016end,codevilla2018end}.

Behavioral cloning has been successfully used for many tasks including off-road driving \cite{muller2006off} and lane following \cite{bojarski2016end}. It is both simple and efficient.
It follows a teacher-student paradigm, where students learn to mimic the teacher's demonstration.
Previous behavioral cloning work for autonomous driving mainly focused on learning only the explicit mapping from the sensory input to the control output, paying no consideration to other potential cues from the teacher that might be beneficial.

While executing tasks, humans attend to behaviorally relevant information using saccades, which direct gaze towards important areas.
In the context of driving, a driver's gaze contains rich information related to his/her intent and decision making.
Researches have shown that novice human learners can benefit from observing experts' gaze \cite{vine2012cheating}.
For example, Yamini \textit{et al.} (\citeyear{yamani2017following}) showed that viewing the expert gaze videos can improve the hazard anticipation ability of novice drivers.
Therefore, it is very promising to investigate whether deep driving networks trained by behavioral cloning might also benefit from exposure to expert gaze patterns.

Whether and how human gaze can help autonomous driving has been under-explored \cite{alletto2016dr}.
Palazzi \textit{et al.} (\citeyear{palazzi2017predicting}) analyzed gaze data in different driving conditions.
They trained a network to predict eye gaze and demonstrated a strong relationship between gaze patterns and driving conditions. However, they did not apply their results to autonomous driving.

We trained a conditional generative adversarial network (GAN) to predict human gaze maps accurately while driving in both familiar and unseen environments. We incorporated the predicted gaze maps into end-to-end networks through two different methods.

First, we added the gaze map as an additional input to the network.
This is a fairly straightforward approach to incorporating additional information.
Treating the gaze map as an additional image input has the disadvantage that it increases the complexity of the network. Since most values of the gaze map are close to zero, this additional complexity is inefficiently utilized.
Second, we used the gaze map to modulate the dropout probability.
Since the human saccadic eye movements directs high resolution processing and attention towards different area of the visual scene, we hypothesized that gaze behavior may be better treated as a modulating effect than as an additional input.

Both approaches improve model generalization to unseen environments. We demonstrate that modelling auxiliary cues not directly related to the control commands improves the performance of imitation learning. This work takes imitation learning to the next step, by showing how deep networks can benefit from a more complete understanding of human expert behaviors.

\section{Methodology}
\label{Methodology}
In this section, we first introduce the network architecture used to estimate the gaze map. Then we describe the two ways we used to incorporate the estimated gaze map into imitation network.

\subsection{Gaze Network}
\begin{figure}[h] 
  \centering
    \includegraphics[width=0.9\columnwidth]{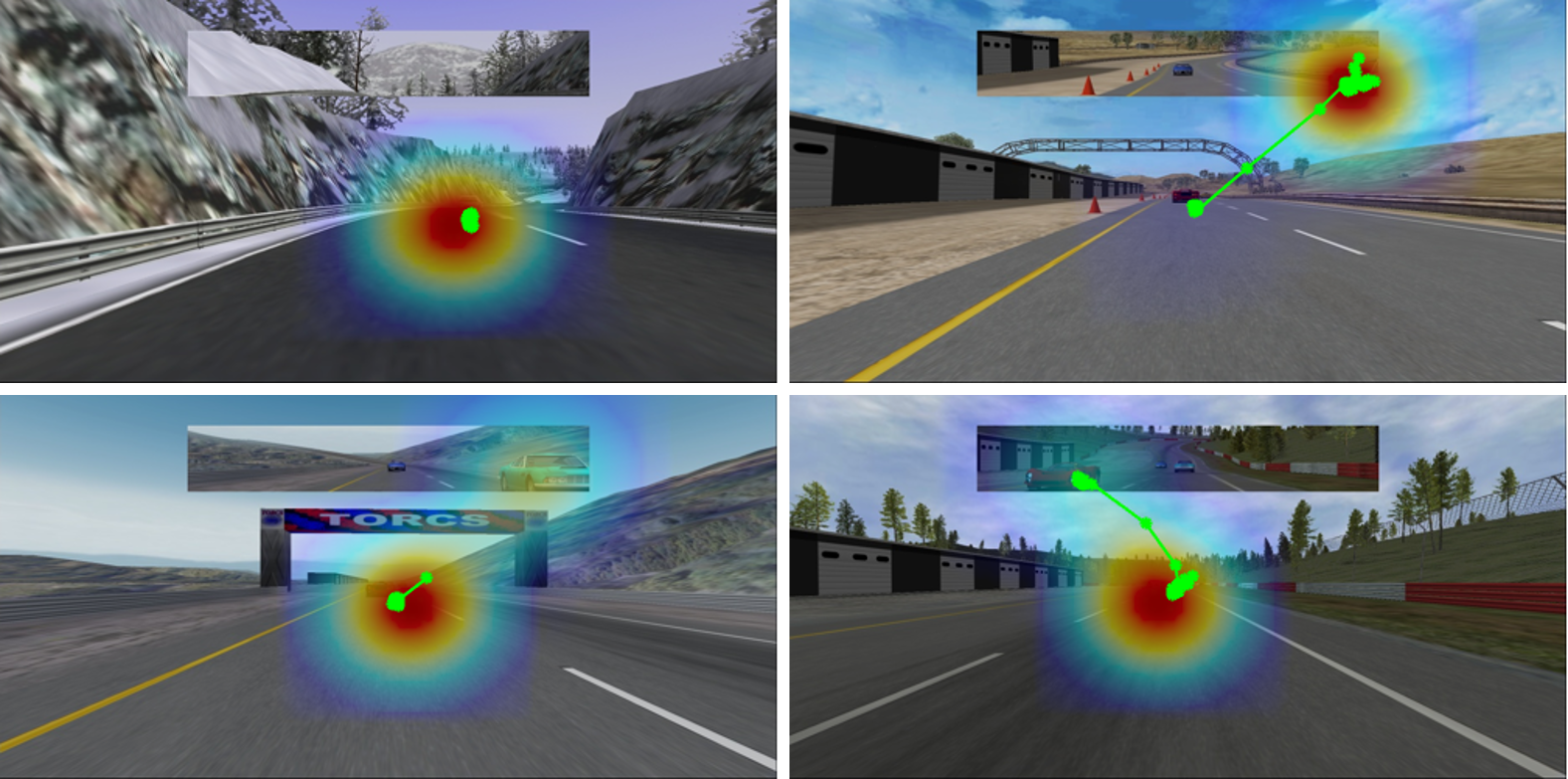}
  \caption{The estimated gaze map and ground truth gaze trajectories visualized as heatmaps and green lines superimposed on the driver-view images. The first row corresponds to the two scenes from environments seen during training. The second row show scenes from two environments unseen during training. On the heatmaps, red areas indicate areas with more fixations.
  } \label{fig:heatmap}
\end{figure}

We generate estimated gaze maps by a conditional GAN following the \textit{Pix2Pix} architecture \cite{isola2017image}. The gaze network was trained on pairs of driver-view images and ground truth gaze maps in a manner similar to the way deep networks have been trained to generate saliency maps \cite{cornia2016deep}.

Based on our observation of the ground truth gaze trajectories, we find the subject mostly looks at the center area of the image. Therefore, we used a static gaze map consisting of a single Gaussian at the center of the image as a baseline for gaze network evaluation.

\subsection{Imitation Learning with Gaze}
\begin{figure}[!ht]
  \centering
    \subfigure[Architecture with gaze map as input]{\includegraphics[width=0.9\columnwidth]{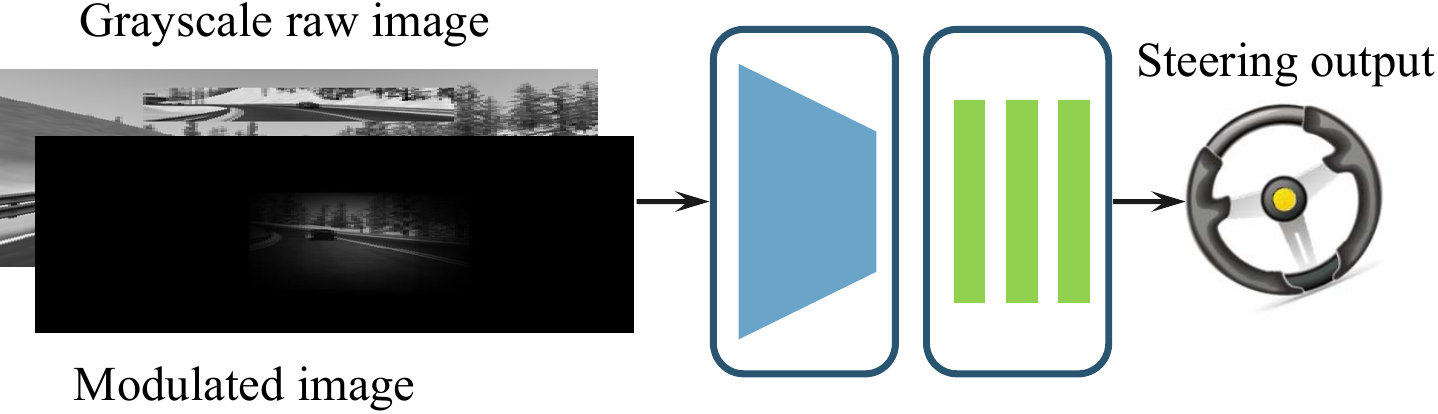}}
    \subfigure[Architecture with gaze map modulated dropout]{\includegraphics[width=0.9\columnwidth]{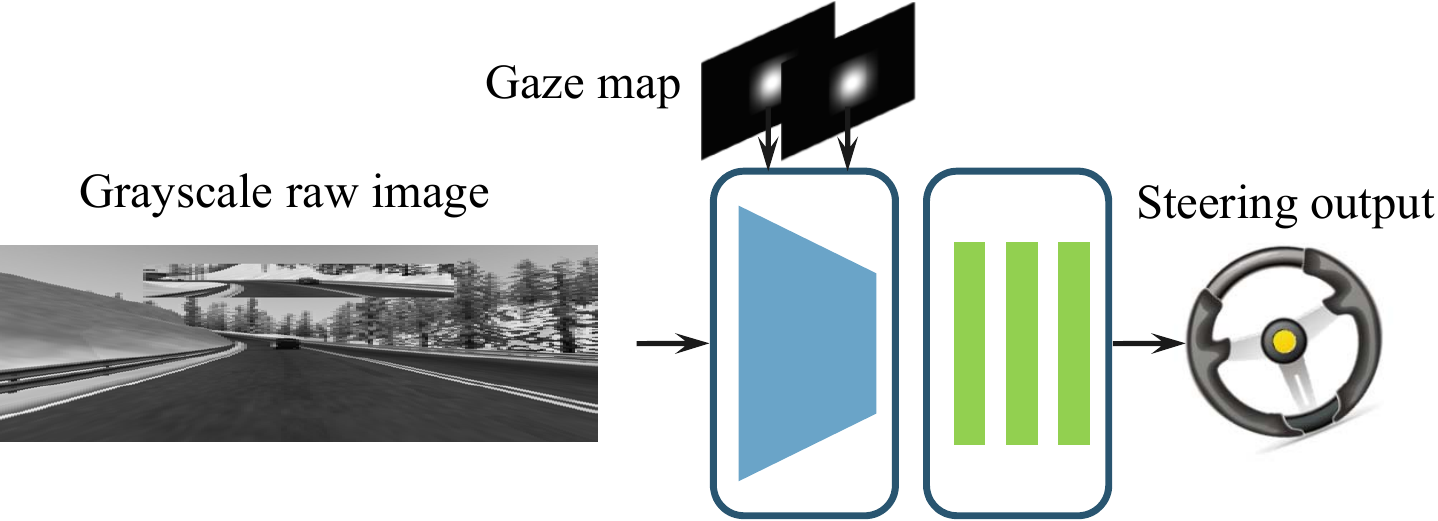}}
  \caption{(a) The gaze map directly modulates the input image by pixel-wise multiplying it. The grayscale image and the modulated image are stacked as the input to the network. (b) The gaze map modulates the dropout probability spatially. The gaze modulated dropout selectively keep the features after first and second convolutional layers.\label{fig:arc}}
\end{figure}

\begin{figure}[!ht] 
  \centering
    \includegraphics[width=0.8\columnwidth]{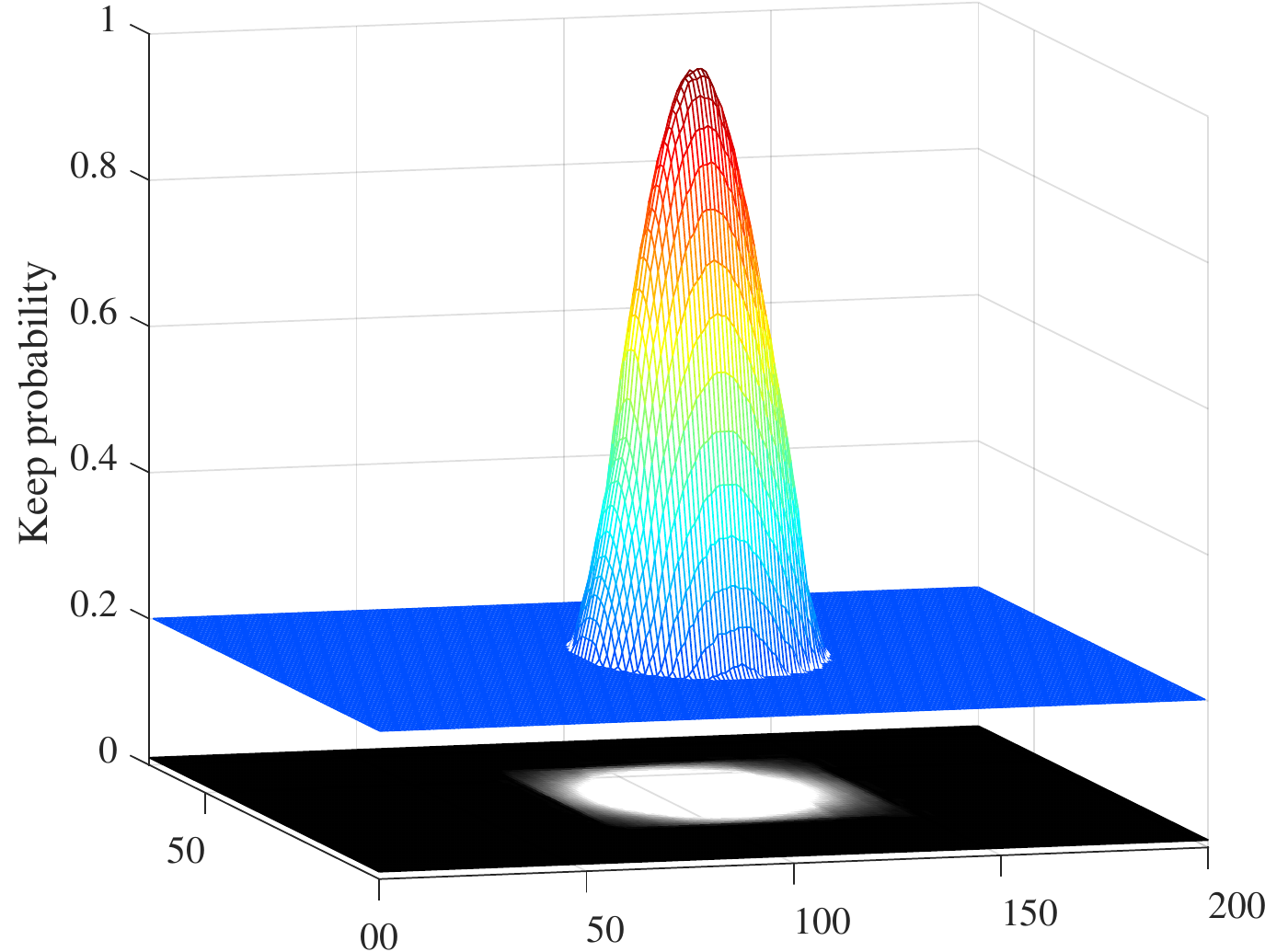}
  \caption{Keep probability settings for gaze-modulated dropout. The bottom image is the corresponding gaze map. At the training stage, keep probabilities are utilized to control whether units are kept or dropped. At the testing state, similar to dropout, the gaze modulated dropout multiplies the features with the keep probability mask generated by the gaze map.} \label{fig:dropout}
\end{figure}

We implemented two different methods to incorporate gaze information into the driving network as shown in Fig.\ref{fig:arc}. Both gaze incorporation methods were implemented and evaluated with the networks based on the PilotNet \cite{bojarski2016end}, which has five convolutional layers and four fully connected layers. The activation function for the networks was the ReLU.

We studied two methods for integrating gaze into the imitation learning network. In the first method, we used the estimated gaze map generated by the gaze network to create an additional input to the network.
As shown in Fig.\ref{fig:arc} (a), we pixel-wise multiplied the greyscale original driver-view image with the estimated gaze map. The modulated and original images were stacked as input to the imitation network.
For more details, please refer to \cite{liu2019gaze}.

In the second method, we used the generated gaze map as the mask to modulate dropout in the network.
Based on the idea that gaze serves as a filter to focus on important areas and de-emphasize task-irrelevant areas, we utilized the gaze to modulate the dropout probability, so that it was higher in uninteresting regions.
The probability setting for the gaze modulated dropout is shown in Fig.\ref{fig:dropout}. As shown in Fig.\ref{fig:arc} (b), we applied the gaze modulated dropout after the first two convolutional layers.
For more details, please refer to \cite{chen2019gaze}.

\begin{table}[!ht]
\begin{tabular}{ccccccc}
\hline
                    &                    & \multicolumn{2}{c}{Seen tracks} & \multicolumn{3}{c}{Unseen tracks} \\ \hline
\multirow{2}{*}{KL} & Estimated gaze & \multicolumn{2}{c}{\textbf{0.69}}        & \multicolumn{3}{c}{\textbf{0.88}}          \\ \cline{2-7}
                    & Central blob       & \multicolumn{2}{c}{2.83}        & \multicolumn{3}{c}{2.32}          \\ \hline
\multirow{2}{*}{CC} & Estimated gaze & \multicolumn{2}{c}{\textbf{0.86}}        & \multicolumn{3}{c}{\textbf{0.83}}          \\ \cline{2-7}
                    & Central blob       & \multicolumn{2}{c}{0.70}        & \multicolumn{3}{c}{0.76}          \\ \hline
\end{tabular}
\caption{The similarity of gaze map estimates to the real gaze map for seen and unseen tracks. KL denotes Kullback-Leibler divergence, and CC denotes Correlation Coefficient.\label{tab:gaze_pred}}
\end{table}

\section{Results}
\label{Results}
\subsection{Gaze Network Evaluation}

The examples of estimated gaze maps superimposed with ground truth gaze trajectories shown in Fig. \ref{fig:heatmap} illustrate the good performance of the gaze map prediction for both seen and unseen environments.

To evaluate the gaze network quantitatively, we compute the two standard metrics for similarity evaluation: the Kullback-Leibler divergence (KL) and the Correlation Coefficient (CC). Smaller KL and larger CC denote better similarity.

As shown in Table.\ref{tab:gaze_pred}, the estimated gaze map closely matched the ground truth gaze map. Compared with the baseline (central gaussian blob), the KL divergence between the estimated gaze map and the real gaze map is significantly smaller on average (75.6\% smaller for seen tracks and 62.1\% smaller for unseen tracks). The CC between estimated gaze map and real gaze map is significantly larger on average (22.9\% larger for seen tracks and 9.2\% larger for unseen tracks).

\subsection{Imitation Network Evaluation}
We trained the imitation network with uniform dropout as the baseline, which we refer to as \textbf{No gaze}.
We considered the cases where the real gaze map, the estimated gaze map and the central Gaussian blob were used as the input to or as dropout modulator for the network.
We refer to the networks with the addition of the input image modulated by the real and estimated gaze maps as \textbf{real/estimated gaze as input} respectively.
We refer to the architecture that modulates dropout with real and estimated gaze map as \textbf{real/estimated gaze dropout} respectively.
The network with the central Gaussian blob (\textbf{Central blob dropout}) tested the effect of simply emphasizing the center region.

\begin{table}[!ht]
\begin{tabular}{lcc}
\hline
\multicolumn{1}{c}{}                                                                   & \begin{tabular}[c]{@{}c@{}}Seen tracks\\ (deg)\end{tabular} & \begin{tabular}[c]{@{}c@{}}Unseen tracks\\ (deg)\end{tabular} \\ \hline
\begin{tabular}[c]{@{}l@{}}No gaze \end{tabular}                    & 2.90                                                  & 5.58                                                   \\
\begin{tabular}[c]{@{}l@{}}Real gaze as input\end{tabular}                                                                     & 2.86                                                  & 4.29                                                   \\
\begin{tabular}[c]{@{}l@{}}Estimated gaze as input\end{tabular}                                                                     & 2.85                                                  & 4.63                                                   \\
\begin{tabular}[c]{@{}l@{}}Real gaze dropout\end{tabular}        & \textbf{2.82}                                                  & \textbf{4.00}                                                   \\
\begin{tabular}[c]{@{}l@{}}Estimated gaze dropout\end{tabular}   & 2.84                                                  & 4.27                                                   \\
\begin{tabular}[c]{@{}l@{}}Central blob dropout\end{tabular} & 2.84                                                  & 4.67                                                   \\ \hline
\end{tabular}
\caption{Average prediction errors between the commands generated by the various models and the human driver.
Errors for the seen tracks are for tracks used in training, but trials not in the training set. Errors for unssen tracks were for tracks and trials not in the training set.
}
\label{tab:error}
\end{table}

\begin{table}[!ht]
\begin{tabular}{llcc}
\hline
                                                                                                                        &                         & W/cars & W/o cars \\ \hline
\multirow{3}{*}{\begin{tabular}[c]{@{}l@{}}Success rate\\ of cars\\ overtaking (\%)\end{tabular}}                       & No gaze         & 67.6      & N/A     \\
                                                                                                                        & Gaze as input & 85.2        & N/A     \\
                                                                                                                        & Gaze dropout  & \textbf{88.9}      & N/A     \\ \hline
\multirow{3}{*}{\begin{tabular}[c]{@{}l@{}}Ave. dist. traveled\\ between two\\infractions (km)\end{tabular}}        & No gaze         & 0.40      & 0.48    \\
                                                                                                                        & Gaze as input & 0.54        & 0.67      \\
                                                                                                                        & Gaze dropout  & \textbf{0.61}      & \textbf{0.79}    \\ \hline
\end{tabular}
\caption{Quantitative performance running on an unseen track in the simulator. Gaze maps are obtained from the gaze network running in real time. We compare the network with estimated gaze-modulated dropout (Gaze dropout), the network with estimated gaze map as input (Gaze as input) and the network with uniform dropout (No gaze). We measure the percentage of successful cars overtaking and the average distance driven between infractions (km). Higher numbers are better for both measurements.}
\label{tab:simulator_test}
\end{table}

\subsubsection{Test on dataset}
We evaluate the average prediction errors between the commands generated by the various models and the human driver. The testing results shown in Table.\ref{tab:error} demonstrate that both approaches decrease the action estimation error in unseen environments.
The network with gaze as input outperforms the baseline (No gaze) by 20.1\% and outperforms the central blob dropout network by 4.5\% for unseen tracks on average.
The network with gaze dropout outperforms the baseline (No gaze) by 25.9\% and outperforms the central blob dropout network by 11.5\% for unseen tracks on average.

The gaze modulated dropout shows better performance than gaze as input.
The network with real gaze dropout outperforms the network with real gaze as input by 6.8\% for unseen tracks and 1.4\% for seen tracks.
The network with estimated gaze dropout outperforms the network with estimated gaze as input by 7.8\% for unseen tracks and 0.35\% for seen tracks.

\subsubsection{Close loop performance}
We further test the performance of the networks running
either "on an unseen track" or "on $X$ unseen tracks" where $X$ is equal to the number of unssen tracks in the simulator.
As the driving networks are intended for autonomous driving applications, the estimated gaze maps from the gaze network are applied in. For each episode, the agent started from a random location and drove along the path, overtaking cars if needed. Human intervention was given when infractions (collisions or drive outside the lane) occur until the car is back on the road. We tested in two cases: track with cars (W/cars) and without cars (W/o cars). In the case without cars, the agent simply follows the road. We evaluated the performance by the success rate in overtaking cars and the average distance traveled between infractions.

The results are summarized in Table \ref{tab:simulator_test}. Both approaches improve the close loop performance. The network with estimated gaze dropout outperforms the network with estimated gaze as input in both cases.
For success rate of overtaking cars, the network with estimated gaze dropout is 31.5\% better than baseline and 4.3\% better than the network with estimated gaze as input.
For the average distance traveled between two infractions averaged over the two test case, the network with estimated gaze dropout is 58.5\% better than the baseline and 15.4\% better than the network with estimated gaze as input.

\section{Conclusions}
\label{Conclusions}
In this paper, we proposed the use of gaze information contained in expert demonstrations to improve the generalization of imitation learning networks for autonomous driving to unseen environments.
We show that a conditional GAN can generate an accurate estimation of human gaze maps during driving.
We studied two ways to incorporate gaze information. Both significantly improve human action estimation accuracy.
Better performance is obtained with the gaze modulated dropout.
This work demonstrates for the first time that it is possible to incorporate human information about gaze behavior into deep driving networks so that they receive similar benefits as obtained when novice human drivers view expert gaze patterns.
This work is an effort to take imitation learning to the next level by implicitly exploiting expert behavior.

\section*{Acknowledgements}
This work was supported by the National Natural Science
Foundation of China (Grant No. U1713211); partially supported
by the HKUST Project IGN16EG12 and Shenzhen
Science, Technology and Innovation Comission (SZSTI)
JCYJ20160428154842603, awarded to Prof. Ming Liu; partially supported
by the Hong Kong Research Grants Council under grant 16213617.

\bibliography{example_paper}
\bibliographystyle{icml2019}


\end{document}